\documentclass[letterpaper, 10pt, conference]{ieeeconf}

\IEEEoverridecommandlockouts
\usepackage{graphicx}
\usepackage{amsmath}
\usepackage{amssymb}
\usepackage{booktabs}
\usepackage{multirow}
\usepackage{cite}
\usepackage{url}
\usepackage{balance}
\usepackage{xcolor}

\title{\LARGE \bf
InfiltrNet: Dual-Branch CNN-Transformer Architecture for\\Brain Tumor Infiltration Risk Prediction
}

\author{S~M~Asif~Hossain and Shruti~Kshirsagar%
\thanks{S~M~Asif~Hossain and Shruti~Kshirsagar are with the School of Computing, Wichita State University, Wichita, Kansas, USA.}}

\begin{document}

\maketitle
\thispagestyle{empty}
\pagestyle{empty}

\begin{abstract}
Gliomas are aggressive brain tumors that infiltrate surrounding tissue beyond the visible tumor margins observed on Magnetic Resonance Imaging (MRI). Predicting the spatial extent of this infiltration is essential for surgical planning and radiation therapy, yet existing deep learning approaches focus on segmenting the visible tumor rather than estimating infiltration risk in the surrounding tissue. This paper presents InfiltrNet, a novel dual-branch architecture that combines a convolutional neural network (CNN) encoder with a Swin Transformer encoder through cross-attention fusion modules to predict three-zone infiltration risk maps from multimodal MRI. A label generation strategy based on distance transforms is proposed to derive reproducible infiltration risk zones from standard Brain Tumor Segmentation (BraTS) annotations. InfiltrNet is trained with a combined Dice-CrossEntropy and boundary-aware loss augmented by auxiliary supervision heads at intermediate decoder levels. Extensive experiments on BraTS 2020 and BraTS 2025 demonstrate that InfiltrNet outperforms five established baselines. Explainability analysis using GradCAM++ and Occlusion sensitivity confirms that the model attends to clinically relevant peritumoral regions.
\end{abstract}

\section{INTRODUCTION}

Gliomas account for approximately 80\% of malignant brain tumors and are characterized by their tendency to infiltrate adjacent brain tissue along white matter tracts and perivascular spaces \cite{louis20212021}. Unlike well-circumscribed tumors, gliomas do not exhibit a clear boundary between malignant and healthy tissue. Tumor cells migrate beyond the contrast-enhancing region visible on conventional MRI, often extending several centimeters into apparently normal brain parenchyma \cite{menze2014multimodal}. This infiltrative behavior poses a fundamental challenge for treatment planning, as insufficient resection leaves behind infiltrating cells that drive tumor recurrence, while excessive resection risks damaging eloquent brain regions \cite{bakas2018identifying , hervey2014role}.

Standard clinical practice relies on T1-weighted contrast-enhanced and T2-FLAIR MRI sequences to identify the enhancing tumor core and surrounding edema. However, histopathological studies have demonstrated that isolated tumor cells are frequently found in tissue that appears normal on MRI, particularly within the peritumoral edema and the adjacent white matter \cite{claes2007diffuse}. The edema region visible on FLAIR sequences contains a mixture of vasogenic fluid and infiltrating tumor cells, and the boundary between tumor-infiltrated tissue and healthy parenchyma remains inherently ambiguous on current imaging modalities \cite{giese2003cost}. This diagnostic uncertainty highlights the clinical need for computational tools that can estimate the spatial distribution of infiltration risk beyond the visible tumor boundary.

Deep learning has achieved remarkable success in brain tumor segmentation through the BraTS challenge \cite{menze2014multimodal, bakas2018identifying}. Architectures such as 3D U-Net \cite{cciccek20163d}, V-Net \cite{milletari2016v}, UNETR \cite{hatamizadeh2022unetr}, Swin-UNETR \cite{hatamizadeh2021swin}, and SegResNet \cite{myronenko20183d} can accurately segment the enhancing tumor, necrotic core, and peritumoral edema from multimodal MRI. However, these models are designed to delineate structures that are already visible on MRI rather than to predict the spatial extent of tumor cell infiltration into surrounding tissue \cite{liu2023deep}. A small number of studies have explored glioma infiltration through biophysical simulations \cite{swanson2000quantitative} and radiomic feature extraction \cite{rathore2018radiomic}, but these approaches require patient-specific parameter estimation and lack integration with end-to-end deep learning frameworks.

In this work, we formulate infiltration risk prediction as a multi-class segmentation task and propose InfiltrNet, a dual-branch architecture that combines local texture analysis with global spatial reasoning through cross-attention fusion. The contributions of this paper are as follows:

\begin{enumerate}
\item A label generation method that derives three-zone infiltration risk maps from standard BraTS segmentation masks using Euclidean distance transforms, providing reproducible training targets for supervised learning.
\item InfiltrNet, a novel architecture that fuses features from a CNN encoder and a Swin Transformer encoder through bidirectional cross-attention modules at every decoder level, capturing both local boundary patterns and global tumor context.
\item Comprehensive evaluation on two independent datasets (BraTS 2020 with 369 patients and BraTS 2025 with 1251 patients) against five established baselines, with ablation studies quantifying the contribution of each component.
\item Explainability analysis using GradCAM++ and Occlusion sensitivity to validate that the model focuses on clinically relevant peritumoral regions.
\end{enumerate}

The remainder of this paper is organized as follows: Section~\ref{sec:relatedwork} reviews related work, Section~\ref{sec:method} describes the proposed methodology, Section~\ref{sec:results} presents the results and discussion, and Section~\ref{sec:conclusion} concludes the paper.

\section{RELATED WORK}
\label{sec:relatedwork}

\subsection{Brain Tumor Segmentation}
The BraTS challenge \cite{menze2014multimodal, bakas2018identifying}
has been the primary benchmark for brain tumor segmentation since 2012.
Early methods such as U-Net \cite{ronneberger2015u} and its 3D extension
\cite{cciccek20163d}, along with V-Net \cite{milletari2016v}, established
strong baselines. Later approaches, including Myronenko
\cite{myronenko20183d}, cascaded U-Net \cite{jiang2019two, karji2024brain},
and nnU-Net \cite{isensee2021nnu}, achieved state-of-the-art performance
through improved architectures and training strategies.

Transformer-based models have since been introduced to capture long-range
dependencies \cite{kshirsagar2022affective, kshirsagar2022quality,
kshirsagar2023task}. UNETR \cite{hatamizadeh2022unetr} replaces CNN
encoders with Vision Transformers \cite{dosovitskiy2020image}, while
Swin-UNETR \cite{hatamizadeh2021swin} uses shifted window attention
\cite{liu2021swin}. TransBTS \cite{wang2021transbts} combines CNN and
transformer features, and self-supervised pretraining further improves
performance \cite{tang2022self}. Despite these advances, existing methods
focus on segmenting visible tumor regions and do not address infiltration
risk beyond tumor boundaries.

\subsection{Tumor Infiltration and Invasion Modeling}
Computational modeling of glioma invasion has been explored primarily through biophysical simulations. Swanson et al. \cite{swanson2000quantitative} modeled tumor cell migration as a reaction-diffusion process governed by spatially varying proliferation and diffusion rates, demonstrating that these rates differ between grey matter and white matter. Jbabdi et al. \cite{jbabdi2005simulation} extended this framework by incorporating patient-specific diffusion tensor imaging data to simulate anisotropic tumor invasion along white matter tracts. Hogea et al. \cite{hogea2008image} proposed a more comprehensive model that couples biomechanical tissue deformation with reaction-diffusion tumor growth dynamics. On the data-driven side, Rathore et al. \cite{rathore2018radiomic} extracted radiomic features from the peritumoral region of multi-parametric MRI to identify distinct infiltration subtypes with prognostic significance. Akbari et al. \cite{akbari2016imaging} applied machine learning classifiers on multi-parametric MRI features to predict the spatial locations of subsequent tumor recurrence in the peritumoral zone.
These approaches face practical limitations for clinical deployment. Explainability methods such as GradCAM++ \cite{chattopadhay2018grad} and Occlusion sensitivity \cite{zeiler2014visualizing} have been widely adopted in medical imaging to validate that deep learning models attend to clinically relevant regions \cite{singh2020explainable}, and we employ both methods in this work. To the best of our knowledge, no prior work has applied modern deep learning segmentation architectures to predict voxel-level infiltration risk zones in the tissue surrounding gliomas.

\section{METHODOLOGY}
\label{sec:method}
In this section, we describe the datasets, label generation  strategy, proposed InfiltrNet architecture, training  strategy, baseline models, and evaluation metrics used 
in this work.
\subsection{Datasets}
Two publicly available datasets from the BraTS challenge are used. BraTS 2020 contains 369 patients with four co-registered MRI modalities (T1, T1ce, T2, FLAIR) at 1~mm isotropic resolution (240$\times$240$\times$155 voxels). Expert annotations define necrotic core (label 1), peritumoral edema (label 2), and enhancing tumor (label 4). The dataset is split into 294 - 37 - 37 for training, validation, and testing, respectively. BraTS 2025 contains 1251 patients with modalities t1c, t1n, t2f, and t2w. Segmentation labels use 1 for necrotic, 2 for edema, and 3 for enhancing. The dataset is split into 1000 - 125 - 126 for training, validation, and testing, respectively.

\subsection{Infiltration Risk Label Generation}
Voxel-level ground truth annotations for tumor infiltration are not available in existing clinical datasets, as obtaining them would require extensive histopathological sampling that is infeasible in routine practice. To address this, we generate infiltration risk labels from the standard BraTS segmentation masks using Euclidean distance transforms. This approach is grounded in the well-established observation that the density of infiltrating glioma cells decreases with increasing distance from the visible tumor margin \cite{swanson2000quantitative, claes2007diffuse}.

Given a BraTS segmentation mask, the tumor core $T_c$ is defined as the union of the necrotic and enhancing sub-regions, the edema region is denoted as $E$, and the complete tumor region is $T = T_c \cup E$. A brain mask $B$ is computed from non-zero voxels in the T2-FLAIR modality. For each brain voxel $v$ outside the tumor core, the Euclidean distance $D(v)$ to the tumor boundary $\partial T$ is computed. The infiltration risk label $L(v)$ is assigned as:

\begin{equation}
L(v) = \begin{cases}
3 & \text{if } v \in E \text{ or } 0 < D(v) \leq 10\text{ mm} \\
2 & \text{if } 10 < D(v) \leq 20\text{ mm} \\
1 & \text{if } D(v) > 20\text{ mm, } v \in B \\
0 & \text{otherwise}
\end{cases}
\end{equation}

Zone 3 (high risk) encompasses the peritumoral edema and tissue within 10~mm of the tumor boundary. Zone 2 (medium risk) covers the 10--20~mm transition region. Zone 1 (low risk) includes brain tissue beyond 20~mm. The tumor core and non-brain regions are assigned label 0.

\subsection{InfiltrNet Architecture}
\begin{figure*}[t]
    \centering
    \includegraphics[width=0.85\textwidth]{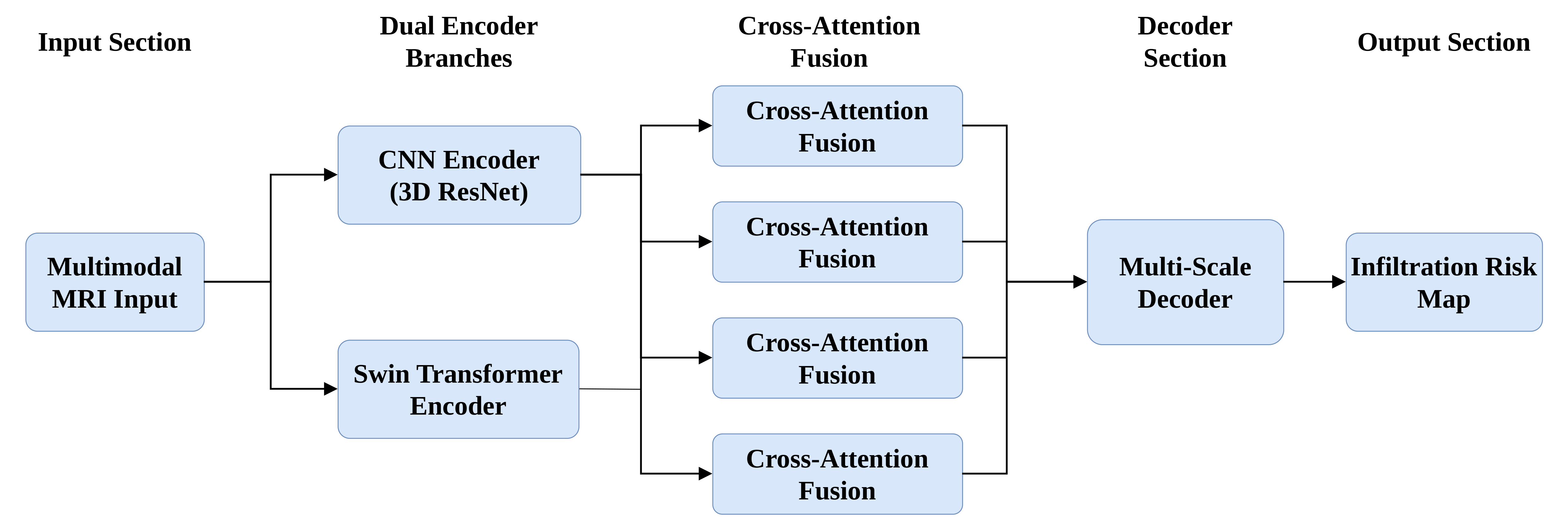}
    \caption{Overview of the InfiltrNet architecture.}
    \label{fig:arch}
\end{figure*}

The proposed architecture consists of two parallel encoder branches, four cross-attention fusion modules, and a multi-scale decoder with auxiliary  heads, as illustrated in Fig.~\ref{fig:arch}.

  \subsubsection{CNN Branch:} A lightweight 3D ResNet encoder is designed to capture local texture patterns and fine-grained boundary features at five resolution levels. Each level contains a residual block comprising two 3$\times$3$\times$3 convolutions with instance normalization and LeakyReLU activation, followed by a residual skip connection. Spatial downsampling between levels is performed using strided convolutions with stride 2. The encoder produces feature maps at resolutions of 1/1, 1/2, 1/4, 1/8, and 1/16 of the input, with channel dimensions of $C$, $C$, $2C$, $4C$, and $8C$, where the base filter count $C$ is set to 32.
   \subsubsection{Swin Transformer Branch:} A Swin Transformer encoder \cite{liu2021swin} is employed to model long-range spatial dependencies through hierarchical shifted window self-attention with feature size $F$=24. The shifted window mechanism computes self-attention within non-overlapping local windows and alternates with shifted partitions to enable cross-window information exchange, providing an efficient alternative to global self-attention. The encoder produces features at resolutions of 1/2, 1/4, 1/8, and 1/16 with channel dimensions of $F$, $2F$, $4F$, and $8F$.
 \subsubsection{Cross-Attention Fusion:} At each decoder level $l$, features from the CNN branch ($\mathbf{C}_l$) and the Swin branch ($\mathbf{S}_l$) are combined through a bidirectional cross-attention module. Both feature maps are first projected to a common channel dimension $d_l$ using 1$\times$1$\times$1 convolutions. The fused representation is computed as:

\begin{equation}
\mathbf{F}_l = \text{Proj}\left[\text{Attn}(\mathbf{Q}_C, \mathbf{K}_S, \mathbf{V}_S) \| \text{Attn}(\mathbf{Q}_S, \mathbf{K}_C, \mathbf{V}_C)\right]
\end{equation}

\noindent where $\text{Attn}(\mathbf{Q}, \mathbf{K}, \mathbf{V}) = \text{softmax}(\mathbf{Q}\mathbf{K}^\top / \sqrt{d_l})\mathbf{V}$ denotes the scaled dot-product attention, $\|$ represents channel-wise concatenation, and $\text{Proj}$ is a 1$\times$1$\times$1 convolution followed by instance normalization and LeakyReLU. 
\subsubsection{Decoder:} The decoder consists of four levels. At each level, the input is upsampled using a transposed convolution with stride 2, concatenated with the corresponding fused skip features $\mathbf{F}_l$, and processed through two 3$\times$3$\times$3 convolutions with instance normalization and LeakyReLU activation. The final output is produced by a 1$\times$1$\times$1 convolution followed by softmax activation, generating a four-class probability map over the entire input volume.

\subsection{Training Strategy}
\subsubsection{Loss Function} The total training loss is a weighted combination of three components:
\begin{equation}
\mathcal{L}_{\text{total}} = \mathcal{L}_{\text{DiceCE}} + \lambda_b \mathcal{L}_{\text{boundary}} + \lambda_a \mathcal{L}_{\text{aux}}
\end{equation}

The Dice-CrossEntropy loss $\mathcal{L}_{\text{DiceCE}}$ jointly optimizes volumetric overlap and voxel-wise classification accuracy with class weights of [0.1, 1.0, 1.5, 2.0] to address the imbalance between background and risk zones. The boundary-aware loss $\mathcal{L}_{\text{boundary}}$ applies an increased penalty to misclassified voxels at zone transitions. A binary boundary mask is computed by comparing each voxel's label with its six immediate spatial neighbors along depth, height, and width. Any voxel for which at least one neighbor carries a different label receives an additional weighting factor of 0.5 in the cross-entropy computation. The coefficients are $\lambda_b = 0.3$ and $\lambda_a = 0.3$.
 \subsubsection{Auxiliary Supervision} To improve gradient flow to the deeper layers of the network, auxiliary prediction heads are attached at three intermediate decoder levels corresponding to 1/2, 1/4, and 1/8 of the input resolution. Each auxiliary head consists of a 1$\times$1$\times$1 convolution that produces four-class predictions at its respective resolution. During training, the ground truth labels are downsampled to match each auxiliary resolution, and the DiceCE loss is computed between the auxiliary predictions and the downsampled labels. The auxiliary loss $\mathcal{L}_{\text{aux}}$ is the average of these three intermediate losses. This mechanism ensures that the cross-attention fusion modules and the deeper encoder layers receive direct supervisory signals during backpropagation, rather than relying solely on gradients that must propagate through the entire decoder path. 

\subsubsection{Data Augmentation and Optimization.} Training patches of 96$^3$ voxels are randomly cropped with balanced positive and negative sampling, with two patches extracted per volume. Data augmentation includes random flips along all three spatial axes and random 90-degree rotations. Each MRI modality is independently normalized using z-score normalization on non-zero voxels. The AdamW optimizer \cite{loshchilov2017decoupled} is used with a learning rate of $10^{-4}$, weight decay of $10^{-5}$, and a cosine annealing schedule with a 5-epoch linear warmup. Mixed-precision training is employed. We utilzed early stopping with patience 15.

\subsubsection{Test-Time Augmentation and Post-Processing.} During inference, sliding window prediction with 50\% overlap is used. Test-time augmentation (TTA) averages softmax predictions across all eight axis-flip combinations. Post-processing removes connected components smaller than 500 voxels and fills holes within each zone. These enhancements are applied exclusively to InfiltrNet.

\subsection{Baseline Models}
Five well-established segmentation architectures are selected as baselines. 3D U-Net \cite{cciccek20163d} is configured with encoder channels of (32, 64, 128, 256, 512) and two residual units per level. V-Net\cite{milletari2016v} employs volumetric 5$\times$5$\times$5 convolutions with its standard configuration. UNETR\cite{hatamizadeh2022unetr} uses a Vision Transformer encoder with 768 hidden dimensions, 12 attention heads, and an MLP dimension of 3072. Swin-UNETR \cite{hatamizadeh2021swin} uses a Swin Transformer encoder with a feature size of 24. SegResNet \cite{myronenko20183d} is configured with 32 initial filters and encoder/decoder block depths of (1,2,2,4) and (1,1,1). All baselines are trained using the same DiceCE loss, AdamW optimizer, learning rate schedule, and data augmentation pipeline as InfiltrNet.

\subsection{Evaluation Metrics}

The InfiltrNet and other baseline models are evaluated using the same set of evaluation metrics to ensure a fair and consistent performance comparison, and those metrics are as follows:
\begin{enumerate}
    \item Dice Similarity Coefficient (DSC): The DSC measures the volumetric overlap between the predicted segmentation $P$ and the ground truth $G$, defined as:
\begin{equation}
\text{DSC} = \frac{2|P \cap G|}{|P| + |G|}
\end{equation}
\item 95th Percentile Hausdorff Distance (HD95): The HD95 measures the 95th percentile of the symmetric surface distances between the boundaries of $P$ and $G$ in millimeters, reflecting boundary accuracy while being robust to small outliers.
\item Intersection over Union (IoU): A stricter overlap measure defined as $|P \cap G| / |P \cup G|$.
\item Volumetric Similarity (VS): Quantifies volume agreement as $1 - |V_P - V_G| / (V_P + V_G)$.

\item Sensitivity: The proportion of ground truth voxels correctly identified, computed as $\text{TP} / (\text{TP} + \text{FN})$.

\item Precision: The proportion of predicted positives that are correct, computed as $\text{TP} / (\text{TP} + \text{FP})$.

\end{enumerate}
All metrics are computed per zone and averaged for mean values.


\section{RESULTS AND DISCUSSION}
\label{sec:results}
In this section, we present and discuss the quantitative  and qualitative experimental results of the proposed InfiltrNet against five established baselines, followed 
by ablation and explainability analyses.
\subsection{Quantitative Evaluation on BraTS 2020}
Table~\ref{tab:brats2020} presents the quantitative comparison on the BraTS 2020 test set comprising 37 patients. InfiltrNet achieves the highest mean Dice score of 0.89, surpassing all five baseline architectures. The improvement is most pronounced in Zone 2 (medium risk), where InfiltrNet achieves a Dice of 0.81 compared to 0.79 for Swin-UNETR, SegResNet, and 3D U-Net. This zone represents the critical transition region where infiltration probability shifts from high to moderate, and accurate delineation of this boundary has direct implications for determining surgical margins and radiation field extent. 
InfiltrNet also achieves the lowest HD95 of 8.90~mm, compared to 9.43~mm for SegResNet and 9.59~mm for 3D U-Net. A lower HD95 indicates fewer large boundary prediction errors, which is clinically significant because even small spatial misplacements at zone transitions can shift the estimated surgical margin by several millimeters. Among the baselines, UNETR exhibits the weakest overall performance with a Dice of 0.86 and HD95 of 12.10~mm, suggesting that the standard ViT encoder lacks the fine-grained spatial precision needed for zone boundary delineation when used without complementary local feature extraction.

\begin{table}[t]
\caption{Quantitative Results on BraTS 2020 Test Set}
\label{tab:brats2020}
\centering
\setlength{\tabcolsep}{2.5pt}
\renewcommand{\arraystretch}{1.1}
\begin{tabular}{l c c c c c c}
\toprule
\textbf{Metric} & \textbf{U-Net} & \textbf{UNETR} & \textbf{Swin} & \textbf{V-Net} & \textbf{SegRes} & \textbf{InfiltrNet} \\
\midrule
Dice (mean) & 0.87 & 0.86 & 0.88 & 0.86 & 0.88 & \textbf{0.89} \\
Dice (Zone 1) & 0.95 & 0.95 & 0.96 & 0.95 & 0.95 & \textbf{0.96} \\
Dice (Zone 2) & 0.79 & 0.76 & 0.79 & 0.78 & 0.79 & \textbf{0.81} \\
Dice (Zone 3) & 0.87 & 0.86 & 0.88 & 0.86 & 0.87 & \textbf{0.89} \\
HD95 (mm) & 9.59 & 12.10 & 9.73 & 9.82 & 9.43 & \textbf{8.90} \\
IoU (mean) & 0.78 & 0.76 & 0.79 & 0.77 & 0.79 & \textbf{0.80} \\
Vol. Similarity & 0.97 & 0.97 & 0.98 & 0.97 & 0.97 & \textbf{0.98} \\
Sensitivity & 0.87 & 0.86 & 0.88 & 0.87 & 0.87 & \textbf{0.89} \\
Precision & 0.86 & 0.85 & 0.87 & 0.85 & 0.87 & \textbf{0.88} \\
\bottomrule
\end{tabular}
\end{table}

\subsection{Quantitative Evaluation on BraTS  2025}
Table~\ref{tab:brats2025} presents the results on BraTS 2025, which contains 1251 patients and provides a more statistically robust evaluation. InfiltrNet achieves a mean Dice of 0.90, representing a 2\% absolute improvement over SegResNet (0.88), the strongest baseline. The performance gap between InfiltrNet and the baselines is wider on this larger dataset compared to BraTS 2020, suggesting that the dual-branch architecture with cross-attention fusion benefits more from additional training data. The cross-attention mechanism requires sufficient training examples to learn effective feature interactions between the CNN and transformer branches.

The HD95 of 10.16~mm represents a 2.22~mm improvement over SegResNet (12.38~mm), indicating substantially more precise boundary prediction. Zone 3 (high risk) shows the largest absolute Dice improvement, with InfiltrNet at 0.91 compared to 0.88 for SegResNet. Accurate prediction of the high-risk zone is the most clinically relevant outcome, as it directly informs surgical resection extent and radiation fie\cite{hervey2014role}. InfiltrNet achieves balanced sensitivity and precision (both 0.91), indicating that it neither over-predicts nor under-predicts the extent of infiltration.

\begin{table}[t]
\caption{Quantitative Results on BraTS 2025 Test Set}
\label{tab:brats2025}
\centering
\setlength{\tabcolsep}{2.5pt}
\renewcommand{\arraystretch}{1.1}
\begin{tabular}{l c c c c c c}
\toprule
\textbf{Metric} & \textbf{U-Net} & \textbf{UNETR} & \textbf{Swin} & \textbf{V-Net} & \textbf{SegRes} & \textbf{InfiltrNet} \\
\midrule
Dice (mean) & 0.86 & 0.83 & 0.85 & 0.85 & 0.88 & \textbf{0.90} \\
Dice (Zone 1) & 0.94 & 0.94 & 0.95 & 0.94 & 0.95 & \textbf{0.97} \\
Dice (Zone 2) & 0.77 & 0.72 & 0.76 & 0.75 & 0.79 & \textbf{0.83} \\
Dice (Zone 3) & 0.86 & 0.82 & 0.84 & 0.86 & 0.88 & \textbf{0.91} \\
HD95 (mm) & 13.09 & 15.78 & 14.38 & 14.68 & 12.38 & \textbf{10.16} \\
IoU (mean) & 0.78 & 0.74 & 0.77 & 0.76 & 0.80 & \textbf{0.84} \\
Vol. Similarity & 0.93 & 0.91 & 0.92 & 0.92 & 0.94 & \textbf{0.97} \\
Sensitivity & 0.86 & 0.83 & 0.85 & 0.85 & 0.88 & \textbf{0.91} \\
Precision & 0.87 & 0.85 & 0.88 & 0.87 & 0.89 & \textbf{0.91} \\
\bottomrule
\end{tabular}
\end{table}

\subsection{Ablation Study}
Table~\ref{tab:ablation} presents an ablation study on BraTS 2025 to quantify the contribution of each component. Removing the boundary-aware loss reduces the mean Dice from 0.90 to 0.87 and increases HD95 from 10.16 to 12.10~mm, confirming its critical role in sharpening zone transitions. Disabling auxiliary supervision decreases the Dice to 0.88, demonstrating the importance of providing direct gradient signals to intermediate decoder and fusion layers during training. Removing TTA and post-processing at test time yields a Dice of 0.89, which still exceeds all baselines, demonstrating that the performance advantage originates primarily from the architecture and training strategy.

The single-branch ablations are particularly informative. Using only the CNN branch produces a Dice of 0.86, comparable to the 3D U-Net baseline \cite{cciccek20163d}, while the Swin branch alone yields 0.85, matching Swin-UNETR \cite{hatamizadeh2021swin}. The full InfiltrNet with cross-attention fusion achieves 0.90, a 4--5\% improvement over either branch in isolation, demonstrating that the fusion mechanism learns complementary feature interactions that cannot be replicated by increasing the capacity of a single encoder.

\begin{table}[t]
\caption{Ablation Study on BraTS 2025}
\label{tab:ablation}
\centering
\setlength{\tabcolsep}{4pt}
\renewcommand{\arraystretch}{1.1}
\begin{tabular}{l c c c c}
\toprule
\textbf{Configuration} & \textbf{Dice} & \textbf{HD95} & \textbf{IoU} & \textbf{Sensitivity} \\
\midrule
InfiltrNet (full) & \textbf{0.90} & \textbf{10.16} & \textbf{0.84} & \textbf{0.91} \\
w/o boundary loss & 0.87 & 12.10 & 0.81 & 0.89 \\
w/o aux. supervision & 0.88 & 11.20 & 0.81 & 0.88 \\
w/o TTA + PP & 0.89 & 10.50 & 0.82 & 0.90 \\
CNN branch only & 0.86 & 13.50 & 0.78 & 0.86 \\
Swin branch only & 0.85 & 14.37 & 0.77 & 0.85 \\
\bottomrule
\end{tabular}
\end{table}

\subsection{Qualitative Infiltration Risk Map Analysis}
\begin{figure}[t]
    \centering
    \includegraphics[width=\columnwidth]{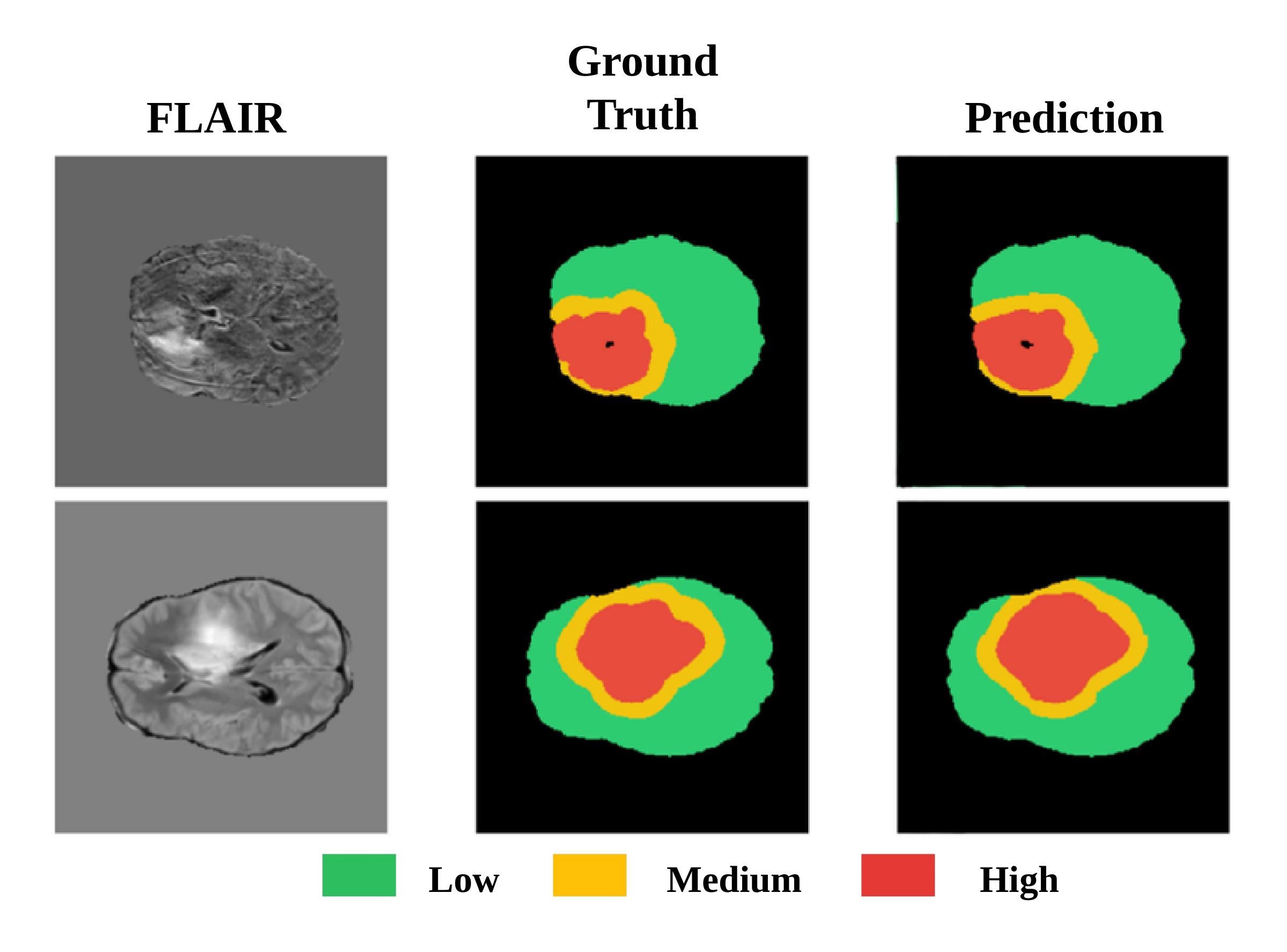}
    \caption{Infiltration risk map predictions for two BraTS 2020 test patients. Each row shows the FLAIR image, ground truth, and InfiltrNet prediction. Green: low risk ($>$20~mm), yellow: medium risk (10--20~mm), red: high risk ($<$10~mm).}
    \label{fig:qualitative}
\end{figure}

Fig.~\ref{fig:qualitative} presents infiltration risk map predictions for two BraTS 2020 test patients. The first row shows a tumor with a prominent necrotic core, where InfiltrNet correctly maintains zone concentricity around the tumor boundary. In the second row, InfiltrNet accurately predicts the concentric risk zones for a large tumor with an irregular boundary, producing smooth transitions between the high-risk, medium-risk, and low-risk regions that closely match the ground truth. In both cases, the predicted zone boundaries closely 
follow the ground truth contours, consistent with the 
quantitative HD95 improvements reported in 
Table~\ref{tab:brats2020}.

\subsection{Explainability and Clinical Relevance Validation}
\begin{figure}[t]
    \centering
    \includegraphics[width=\columnwidth]{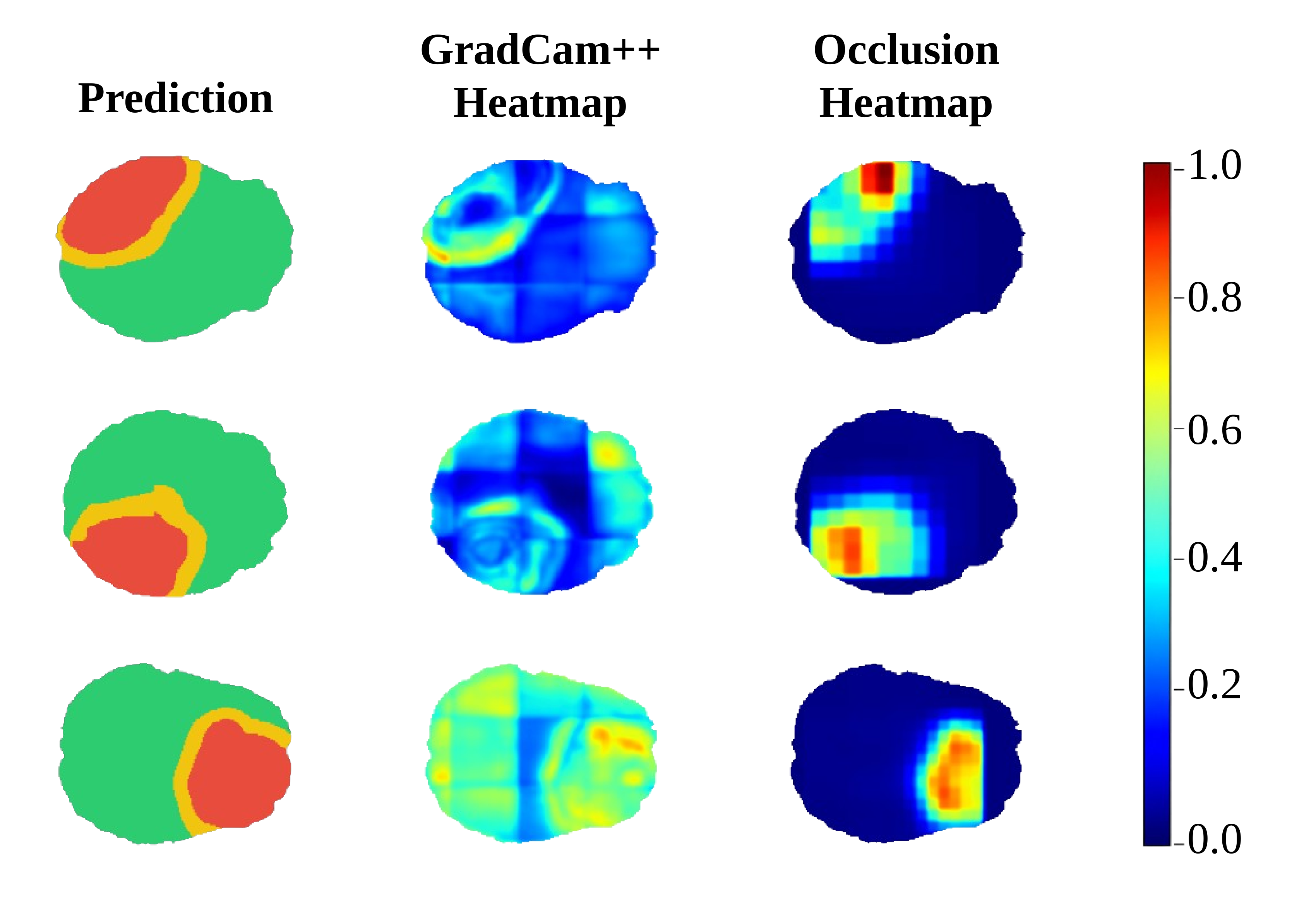}
    \caption{Explainability analysis for three test patients. Each row shows the prediction, GradCAM++ overlay and heatmap, and Occlusion overlay and heatmap.}
    \label{fig:xai}
\end{figure}

Fig.~\ref{fig:xai} presents the explainability analysis for three test patients using GradCAM++ and Occlusion sensitivity\cite{chattopadhay2018grad, zeiler2014visualizing}. The GradCAM++ heatmap columns reveal that InfiltrNet concentrates its feature-level attention at the peritumoral boundary, particularly in regions where the high-risk zone transitions into the medium-risk zone. This attention pattern is visible as warm colors at the tumor periphery in the overlay images and as localized high-intensity regions in the standalone heatmaps. The attention is not uniformly distributed across the brain but is specifically focused on the tissue immediately surrounding the tumor, consistent with the significance of this region for infiltration assessment\cite{claes2007diffuse, 
giese2003cost}. The Occlusion sensitivity maps provide independent 
confirmation of these findings through a fundamentally 
different methodology. By systematically masking input regions 
with multi-scale patches and measuring the resulting prediction 
change, the Occlusion analysis identifies which input regions 
most strongly influence the model output. The  
heatmap columns show that masking the peritumoral boundary 
region produces the largest confidence drop, while masking 
distant brain tissue has minimal effect. The spatial agreement 
between GradCAM++ and Occlusion sensitivity across all three 
patients provides strong evidence that InfiltrNet has learned 
to focus on tissue features relevant to infiltration risk 
rather than relying on dataset artifacts or spurious image 
patterns  a validation criterion that has been shown 
critical for establishing clinical trustworthiness of deep 
learning models in medical 
imaging~\cite{singh2020explainable}.

\section{CONCLUSION}
\label{sec:conclusion}

This paper presented InfiltrNet, a dual-branch architecture for predicting brain tumor infiltration risk zones from multimodal MRI. A label generation method based on Euclidean distance transforms was proposed to derive three-zone risk maps from standard BraTS segmentation annotations. InfiltrNet combines a CNN encoder for local texture analysis with a Swin Transformer encoder for global spatial reasoning, connected through bidirectional cross-attention fusion at every decoder level. Experiments on BraTS 2020 and BraTS 2025 demonstrated consistent improvements over five established architectures, and ablation studies confirmed that each component contributes meaningfully to the overall performance. Explainability analysis using two independent methods validated that the model attends to clinically relevant peritumoral regions. Future work will focus on integrating diffusion tensor imaging to capture directional infiltration along white matter pathways.

\balance

\bibliographystyle{IEEEtran}
\bibliography{references}

@article{louis20212021,
  title={The 2021 WHO classification of tumors of the central nervous system: a summary},
  author={Louis, David N and Perry, Arie and Wesseling, Pieter and Brat, Daniel J and others},
  journal={Neuro-oncology},
  volume={23},
  number={8},
  pages={1231--1251},
  year={2021},
  publisher={Oxford University Press US}
}

@article{kshirsagar2022quality,
  title={Quality-aware bag of modulation spectrum features for robust speech emotion recognition},
  author={Kshirsagar, Shruti Rajendra and Falk, Tiago Henrik},
  journal={IEEE Transactions on Affective Computing},
  volume={13},
  number={4},
  pages={1892--1905},
  year={2022},
  publisher={IEEE}
}

@article{kshirsagar2023task,
  title={Task-specific speech enhancement and data augmentation for improved multimodal emotion recognition under noisy conditions},
  author={Kshirsagar, Shruti and Pendyala, Anurag and Falk, Tiago H},
  journal={Frontiers in Computer Science},
  volume={5},
  pages={1039261},
  year={2023},
  publisher={Frontiers Media SA}
}

@book{kshirsagar2022affective,
  title={Affective Human-Machine Interfaces: Towards Multi-lingual, Environment-Robust Emotion Detection from Speech},
  author={Kshirsagar, Shruti Rajendra},
  year={2022},
  publisher={Institut National de la Recherche Scientifique (Canada)}
}

@mastersthesis{karji2024brain,
  author       = {Karji, Fatemeh},
  title        = {Brain Tumor Segmentation Using Deep Learning Techniques 
                  on Multi-Institutional {MRI} Datasets},
  school       = {School of Computing, Wichita State University},
  address      = {Wichita, KS, USA},
  year         = {2024},
  month        = {December},
  type         = {M.S. Thesis},
  url          = {https://hdl.handle.net/10057/29112}
}

@article{menze2014multimodal,
  title={The multimodal brain tumor image segmentation benchmark (BRATS)},
  author={Menze, Bjoern H and Jakab, Andras and Bauer, Stefan and Kalpathy-Cramer, Jayashree and others},
  journal={IEEE transactions on medical imaging},
  volume={34},
  number={10},
  pages={1993--2024},
  year={2014},
  publisher={IEEE}
}

@article{bakas2018identifying,
  title={Identifying the best machine learning algorithms for brain tumor segmentation, progression assessment, and overall survival prediction in the BRATS challenge},
  author={Bakas, Spyridon and Reyes, Mauricio and Jakab, Andras and Bauer, Stefan and others},
  journal={arXiv preprint arXiv:1811.02629},
  year={2018}
}

@article{hervey2014role,
  title={Role of surgical resection in low-and high-grade gliomas},
  author={Hervey-Jumper, Shawn L and Berger, Mitchel S},
  journal={Current treatment options in neurology},
  volume={16},
  number={4},
  pages={284},
  year={2014},
  publisher={Springer}
}

@article{claes2007diffuse,
  title={Diffuse glioma growth: a guerilla war},
  author={Claes, An and Idema, Albert J and Wesseling, Pieter},
  journal={Acta neuropathologica},
  volume={114},
  number={5},
  pages={443--458},
  publisher={Springer}
}

@article{giese2003cost,
  title={Cost of migration: invasion of malignant gliomas and implications for treatment},
  author={Giese, Arie and Bjerkvig, Rolf and Berens, Michael E and Westphal, Manfred},
  journal={Journal of clinical oncology},
  volume={21},
  number={8},
  pages={1624--1636},
  year={2003},
  publisher={American Society of Clinical Oncology}
}

@article{isensee2021nnu,
  title={nnU-Net: a self-configuring method for deep learning-based biomedical image segmentation},
  author={Isensee, Fabian and Jaeger, Paul F and Kohl, Simon AA and Petersen, Jens and Maier-Hein, Klaus H},
  journal={Nature methods},
  volume={18},
  number={2},
  year={2021},
  publisher={Nature Publishing Group US New York}
}

@article{liu2023deep,
  title={Deep learning based brain tumor segmentation: a survey},
  author={Liu, Zhihua and Tong, Lei and Chen, Long and Jiang, Zheheng and others},
  journal={Complex \& intelligent systems},
  volume={9},
  number={1},
  pages={1001--1026},
  year={2023},
  publisher={Springer}
}

@inproceedings{ronneberger2015u,
  title={U-net: Convolutional networks for biomedical image segmentation},
  author={Ronneberger, Olaf and Fischer, Philipp and Brox, Thomas},
  booktitle={International Conference on Medical image computing and computer-assisted intervention},
  pages={234--241},
  year={2015},
  organization={Springer}
}

@inproceedings{cciccek20163d,
  title={3D U-Net: learning dense volumetric segmentation from sparse annotation},
  author={{\c{C}}i{\c{c}}ek, {\"O}zg{\"u}n and Abdulkadir, Ahmed and Lienkamp, Soeren S and Brox, Thomas and Ronneberger, Olaf},
  booktitle={International conference on medical image computing and computer-assisted intervention},
  year={2016},
  organization={Springer}
}

@inproceedings{milletari2016v,
  title={V-net: Fully convolutional neural networks for volumetric medical image segmentation},
  author={Milletari, Fausto and Navab, Nassir and Ahmadi, Seyed-Ahmad},
  booktitle={2016 fourth international conference on 3D vision (3DV)},
  year={2016},
  organization={Ieee}
}

@inproceedings{hatamizadeh2022unetr,
  title={Unetr: Transformers for 3d medical image segmentation},
  author={Hatamizadeh, Ali and Tang, Yucheng and Nath, Vishwesh and Yang, Dong and others},
  booktitle={Proceedings of the IEEE/CVF winter conference on applications of computer vision},
  pages={574--584},
  year={2022}
}

@inproceedings{hatamizadeh2021swin,
  title={Swin unetr: Swin transformers for semantic segmentation of brain tumors in mri images},
  author={Hatamizadeh, Ali and Nath, Vishwesh and Tang, Yucheng and Yang, Dong and others},
  booktitle={International MICCAI brainlesion workshop},
  pages={272--284},
  year={2021},
  organization={Springer}
}

@inproceedings{myronenko20183d,
  title={3D MRI brain tumor segmentation using autoencoder regularization},
  author={Myronenko, Andriy},
  booktitle={International MICCAI brainlesion workshop},
  pages={311--320},
  year={2018},
  organization={Springer}
}

@inproceedings{jiang2019two,
  title={Two-stage cascaded u-net: 1st place solution to brats challenge 2019 segmentation task},
  author={Jiang, Zeyu and Ding, Changxing and Liu, Minfeng and Tao, Dacheng},
  booktitle={International MICCAI brainlesion workshop},
  pages={231--241},
  year={2019},
  organization={Springer}
}

@article{swanson2000quantitative,
  title={A quantitative model for differential motility of gliomas in grey and white matter},
  author={Swanson, Kristin R and Alvord Jr, Ellsworth C and Murray, James D},
  journal={Cell proliferation},
  volume={33},
  number={5},
  pages={317--329},
  year={2000},
  publisher={Wiley Online Library}
}

@article{rathore2018radiomic,
  title={Radiomic MRI signature reveals three distinct subtypes of glioblastoma with different clinical and molecular characteristics, offering prognostic value beyond IDH1},
  author={Rathore, Saima and Akbari, Hamed and Rozycki, Martin and Abdullah, Kalil G and others},
  journal={Scientific reports},
  volume={8},
  number={1},
  pages={5087},
  year={2018},
  publisher={Nature Publishing Group UK London}
}

@article{jbabdi2005simulation,
  title={Simulation of anisotropic growth of low-grade gliomas using diffusion tensor imaging},
  author={Jbabdi, Sa{\^a}d and Mandonnet, Emmanuel and Duffau, Hugues and Capelle, Laurent and others},
  journal={Magnetic Resonance in Medicine: An Official Journal of the International Society for Magnetic Resonance in Medicine},
  volume={54},
  number={3},
  pages={616--624},
  year={2005},
  publisher={Wiley Online Library}
}

@article{hogea2008image,
  title={An image-driven parameter estimation problem for a reaction--diffusion glioma growth model with mass effects},
  author={Hogea, Cosmina and Davatzikos, Christos and Biros, George},
  journal={Journal of mathematical biology},
  volume={56},
  number={6},
  pages={793--825},
  year={2008},
  publisher={Springer}
}

@article{akbari2016imaging,
  title={Imaging surrogates of infiltration obtained via multiparametric imaging pattern analysis predict subsequent location of recurrence of glioblastoma},
  author={Akbari, Hamed and Macyszyn, Luke and Da, Xiao and Bilello, Michel and others},
  journal={Neurosurgery},
  volume={78},
  number={4},
  pages={572--580},
  year={2016},
  publisher={LWW}
}

@inproceedings{wang2021transbts,
  title={Transbts: Multimodal brain tumor segmentation using transformer},
  author={Wang, Wenxuan and Chen, Chen and Ding, Meng and Yu, Hong and others},
  booktitle={International conference on medical image computing and computer-assisted intervention},
  pages={109--119},
  year={2021},
  organization={Springer}
}

@article{dosovitskiy2020image,
  title={An image is worth 16x16 words: Transformers for image recognition at scale},
  author={Dosovitskiy, Alexey and Beyer, Lucas and Kolesnikov, Alexander and Weissenborn, Dirk and others},
  journal={arXiv preprint arXiv:2010.11929},
  year={2020}
}

@inproceedings{liu2021swin,
  title={Swin transformer: Hierarchical vision transformer using shifted windows},
  author={Liu, Ze and Lin, Yutong and Cao, Yue and Hu, Han and others},
  booktitle={Proceedings of the IEEE/CVF international conference on computer vision},
  year={2021}
}

@inproceedings{chattopadhay2018grad,
  title={Grad-cam++: Generalized gradient-based visual explanations for deep convolutional networks},
  author={Chattopadhay, Aditya and Sarkar, Anirban and Howlader, Prantik and Balasubramanian, Vineeth N},
  booktitle={2018 IEEE winter conference on applications of computer vision (WACV)},
  pages={839--847},
  year={2018},
  organization={IEEE}
}

@inproceedings{zeiler2014visualizing,
  title={Visualizing and understanding convolutional networks},
  author={Zeiler, Matthew D and Fergus, Rob},
  booktitle={European conference on computer vision},
  pages={818--833},
  year={2014},
  organization={Springer}
}

@article{singh2020explainable,
  title={Explainable deep learning models in medical image analysis},
  author={Singh, Amitojdeep and Sengupta, Sourya and Lakshminarayanan, Vasudevan},
  journal={Journal of imaging},
  volume={6}
}

@inproceedings{tang2022self,
  title={Self-supervised pre-training of swin transformers for 3d medical image analysis},
  author={Tang, Yucheng and Yang, Dong and Li, Wenqi and Roth, Holger R and others},
  booktitle={Proceedings of the IEEE/CVF conference on computer vision and pattern recognition},
  pages={20730--20740},
  year={2022}
}

@article{loshchilov2017decoupled,
  title={Decoupled weight decay regularization},
  author={Loshchilov, Ilya and Hutter, Frank},
  journal={arXiv preprint arXiv:1711.05101},
  year={2017}
}

\end{document}